# Lifted Relax, Compensate and then Recover: From Approximate to Exact Lifted Probabilistic Inference

**Guy Van den Broeck***
Department of Computer Science
KU Leuven
guy.vandenbroeck@cs.kuleuven.be

**Arthur Choi** and **Adnan Darwiche**
Computer Science Department
University of California, Los Angeles
{aychoi,darwiche}@cs.ucla.edu

## Abstract

We propose an approach to lifted approximate inference for first-order probabilistic models, such as Markov logic networks. It is based on performing exact lifted inference in a simplified first-order model, which is found by relaxing first-order constraints, and then compensating for the relaxation. These simplified models can be incrementally improved by carefully recovering constraints that have been relaxed, also at the first-order level. This leads to a spectrum of approximations, with lifted belief propagation on one end, and exact lifted inference on the other. We discuss how relaxation, compensation, and recovery can be performed, all at the first-order level, and show empirically that our approach substantially improves on the approximations of both propositional solvers and lifted belief propagation.

## 1 INTRODUCTION

Probabilistic logic models combine aspects of first-order logic and probabilistic graphical models, enabling them to model complex logical and probabilistic interactions between large numbers of objects (Getoor and Taskar, 2007; De Raedt et al., 2008). This level of expressivity comes at the cost of increased complexity of inference, motivating a new line of research in lifted inference algorithms (Poole, 2003). These algorithms exploit logical structure and symmetries in probabilistic logics to perform efficient inference in these models.

For exact inference, lifted variants of variable elimination (Poole, 2003; de Salvo Braz et al., 2005; Milch et al., 2008; Taghipour et al., 2012) and weighted model counting (Gogate and Domingos, 2011; Van den Broeck et al., 2011) have been proposed. With some exceptions, lifted approximate inference has focused on lifting iterative belief propagation (Singla and Domingos, 2008; Kersting et al., 2009).

In this paper, we propose an approach to *approximate* lifted inference that is based on performing *exact* lifted inference in a simplified first-order model. Namely, we simplify the structure of a first-order model until it is amenable to exact lifted inference, by relaxing first-order equivalence constraints in the model. Relaxing equivalence constraints ignores (many) dependencies between random variables, so we compensate for this relaxation by restoring a weaker notion of equivalence, in a lifted way. We then incrementally improve this approximation by recovering first-order equivalence constraints back into the model.

In fact our proposal corresponds to an approach to approximate inference, called Relax, Compensate and then Recover (RCR) for (ground) probabilistic graphical models.[1] For such models, the RCR framework gives rise to a spectrum of approximations, with iterative belief propagation on one end (when we use the coarsest possible model), and exact inference on the other (when we use the original model). In this paper, we show how relaxations, compensations and recovery can all be performed at the first-order level, giving rise to a spectrum of first-order approximations, with lifted first-order belief propagation on one end, and exact lifted inference in the other.

We evaluate our approach on benchmarks from the lifted inference literature. Experiments indicate that recovering a small number of first-order equivalences can improve on the approximations of lifted belief propagation by several orders of magnitude. We show that, compared to Ground RCR, Lifted RCR can re-cover many more equivalences in the same amount of time, leading to better approximations.

---

*Part of this research was conducted while the author was a visiting student at UCLA.

[1] A solver based on the RCR framework won first place in two categories evaluated at the UAI'10 approximate inference challenge (Elidan and Globerson, 2010).

## 2 RCR FOR GROUND MLNs

While our main objective is to present a lifted version of the RCR framework, we start by adapting RCR to ground Markov logic networks (MLNs). This is meant to both motivate the specifics of Lifted RCR and to provide a basis for its semantics (i.e., the correctness of Lifted RCR will be against Ground RCR). RCR can be understood in terms of three steps: Relaxation (R), Compensation (C), and Recovery (R). Next, we introduce MLNs and examine each of these steps.

### 2.1 MARKOV LOGIC NETWORKS

We first introduce some standard concepts from function-free first-order logic. An *atom* $p(t_1, \ldots, t_n)$ consists of a predicate $p/n$ of arity $n$ followed by $n$ arguments, which are either (lowercase) *constants* or (uppercase) *logical variables*. A formula combines atoms with logical connectives (e.g., $\wedge$, $\Leftrightarrow$). A formula is *ground* if it does not contain any logical variables. The groundings of a formula are the formulas obtained by substituting all variables for constants.

Many probabilistic logical languages have been proposed in recent years. We will work with one such language: Markov logic networks (MLN) (Richardson and Domingos, 2006). An MLN is a set of tuples $(w, f)$, where $w$ is a real number representing a weight and $f$ is a formula in function-free first-order logic. First-order logic formulas without a weight are called *hard formulas* and correspond to formulas with an infinite weight. We will assume that all logical variables in $f$ are universally quantified.[2] Consider the MLN

$$1.3 \quad \text{smokes}(X) \Rightarrow \text{cancer}(X) \qquad (1)$$
$$1.5 \quad \text{smokes}(X) \wedge \text{friends}(X, Y) \Rightarrow \text{smokes}(Y) \qquad (2)$$

which states that (1) smokers are more likely to get cancer and (2) smokers are more likely to be friends with other smokers (Singla and Domingos, 2008). This MLN will be a running example throughout this paper.

The *grounding* of an MLN $\Delta$ is the MLN which results from replacing each formula in $\Delta$ with all its groundings (using the same weight). For the domain $\{a, b\}$, the above first-order MLN represents the following ground MLN:

$$1.3 \quad \text{smokes}(a) \Rightarrow \text{cancer}(a) \qquad (3)$$
$$1.3 \quad \text{smokes}(b) \Rightarrow \text{cancer}(b)$$
$$1.5 \quad \text{smokes}(a) \wedge \text{friends}(a, a) \Rightarrow \text{smokes}(a)$$
$$1.5 \quad \text{smokes}(a) \wedge \text{friends}(a, b) \Rightarrow \text{smokes}(b)$$
$$1.5 \quad \text{smokes}(b) \wedge \text{friends}(b, a) \Rightarrow \text{smokes}(a)$$
$$1.5 \quad \text{smokes}(b) \wedge \text{friends}(b, b) \Rightarrow \text{smokes}(b)$$

This ground MLN contains eight different random variables, which correspond to all groundings of atoms smokes($X$), cancer($X$) and friends($X, Y$). This leads to a distribution over $2^8$ possible worlds. The weight of each world is simply the product of all weights $e^w$, where $(w, f)$ is a ground MLN formula and $f$ is satisfied by the world. The weights of worlds that do not satisfy a hard formula are set to zero. The probabilities of worlds are obtained by normalizing their weights. The *ground distribution* of $\Delta$ is the distribution induced by the grounding of $\Delta$.

### 2.2 GROUND RELAXATION

Relaxation is the process of ignoring interactions between the formulas of a ground MLN. An interaction takes place when the same ground atom $a_g$ appears in more than one ground formula in an MLN. We can ignore this interaction via a two step process. First, we rename one occurrence of $a_g$ into, say, $a'_g$, through a process that we call *cloning*. We then assert an equivalence constraint between the original ground atom $a_g$ and its clone, $a_g \Leftrightarrow a'_g$. At this point, we can ignore the interaction by simply dropping the equivalence constraint, through a process that we call *relaxation*. Bringing back the equivalence is known as *recovery* and will be discussed in more detail later.

The smokes($a$) atom in Formula 3 leads to an interaction between this formula and some of the other five formulas in the ground MLN. To ignore this interaction, we first rename this atom occurrence into smokes$_1$($a$) leading to the modified formula

$$1.3 \quad \text{smokes}_1(a) \Rightarrow \text{cancer}(a) \qquad (4)$$

which replaces Formula 3 in the MLN. The corresponding equivalence constraint is

$$\text{smokes}_1(a) \Leftrightarrow \text{smokes}(a) \qquad (5)$$

Dropping this constraint amounts to removing the interaction between Formula 4 and the rest of the MLN.

### 2.3 GROUND COMPENSATION

When relaxing a constraint $a_g \Leftrightarrow a'_g$, we ignore a connection between the ground atoms $a_g$ and $a'_g$. We can *compensate* for this loss by adding two weighted atoms

$$w : \quad a_g \qquad \text{and} \qquad w' : \quad a'_g$$

If the weights $w$ and $w'$ are chosen carefully, one can reestablish a weaker connection between the ground atoms. For example, one can choose these weights to ensure that the ground atoms have the same probability, establishing a weaker notion of equivalence.

---
[2] We transform existential quantifiers into disjunctions.

We will now suggest a specific compensation scheme based on a key result from Choi and Darwiche (2006). Suppose that we relax a single equivalence constraint, $a_g \Leftrightarrow a'_g$, which splits the MLN into two disconnected components, one containing atom $a_g$ and another containing atom $a'_g$. Suppose further that we choose the compensations $w$ and $w'$ such that

$$\Pr(a_g) = \Pr(a'_g) = \frac{e^{w+w'}}{1+e^{w+w'}}. \quad (6)$$

We now have a number of guarantees. First, the resulting MLN will yield exact results when computing the probability of any ground atom. Second, the compensations $w$ and $w'$ can be identified by finding a fixed point for the following equations:

$$w_{i+1} = \log\left(\Pr{}_i(a'_g)\right) - \log\left(\Pr{}_i(\neg a'_g)\right) - w'_i$$
$$w'_{i+1} = \log\left(\Pr{}_i(a_g)\right) - \log\left(\Pr{}_i(\neg a_g)\right) - w_i. \quad (7)$$

Following Choi and Darwiche (2006), we will seek compensations using these update equations even when the relaxed equivalence constraint does not disconnect the MLN, and even when relaxing multiple equivalence constraints. In this more general case, a fixed-point to the above equations will still guarantee the weak equivalence given in (6). However, when computing the probabilities of ground atoms, we will only get approximations instead of exact results.[3]

Searching for compensations using Equations 7 will lead to the generation of a sequence of MLNs that differ only on the weights of atoms added during the compensation process. The first MLN in this sequence is obtained by using zero weights for all compensating atoms, leading to an initial ground distribution $\Pr_0$. Each application of Equations 7 will then lead to a new MLN (with new compensations) and, hence, a new ground distribution, $\Pr_{i+1}$. Upon convergence, the resulting MLN and its ground distribution will then be used for answering queries. This is typically done using an exact inference algorithm as one usually relaxes enough equivalence constraints to make the ground MLN amenable to exact inference. Note that applying Equations 7 also requires exact inference, as one must compute the probabilities $\Pr_i(a_g)$ and $\Pr_i(a'_g)$.

### 2.4 GROUND RECOVERY

Now that we can relax equivalences and compensate for their loss, the remaining question is which equivalences to relax. In general, deciding which equivalences to relax is hard, because it requires inference in the original model, which is intractable. Instead, Choi and Darwiche (2006) take the approach of relaxing every equivalence constraint and then incrementally recovering them as time and exact inference allow.

It follows from their results that when (i) relaxing all equivalence constraints, (ii) using the above compensation scheme and (iii) doing exact inference in the approximate model, the approximate marginals found correspond to the approximations found by *iterative belief propagation* (IBP) (Pearl, 1988). The connection to IBP is even stronger: the compensating weights computed in each iteration of Equations 7 exactly correspond to the messages passed by IBP.

Several heuristics have been proposed to decide which equivalences to recover, by doing inference in the relaxed model. We will work with the *residual recovery* heuristic (Choi and Darwiche, 2011). It is based on the practical observation that when IBP converges easily, the quality of its approximation is high. The heuristic tries to recover those constraints that have the most difficulty converging throughout the iterative process, i.e., those that least satisfy Equation 6. We measure this by keeping track of the three-way symmetric KL divergence between the three terms of Equation 6.

## 3 LIFTED RCR

We now introduce a lifted version of the relax, compensate and recover framework, which is meant to operate directly on first-order MLNs without necessarily having to ground them. Lifted RCR is based on first-order relaxation, compensation and recovery.

### 3.1 FIRST-ORDER RELAXATION

We will now discuss a first-order notion of relaxation where the goal is to ignore interactions between ground MLN formulas, yet without necessarily having to fully ground the MLN. This requires a first-order version of atom cloning and first-order equivalences.

**Definition 1** (First-Order Cloning). Cloning an atom occurrence in an MLN formula amounts to renaming the atom by concatenating its predicate with (i) an identifier of the formula, (ii) an identifier of the occurrence of the atom within the formula, and (iii) the logical variables appearing in the atom's formula.

For example, the first-order cloning of the atom occurrence smokes($Y$) in Formula 2 gives

$$1.5 \quad \text{smokes}(X) \wedge \text{friends}(X,Y)$$
$$\Rightarrow \text{smokes}_{2b<X,Y>}(Y) \quad (8)$$

---
[3] In this more general case, there is no longer a guarantee that Equations 7 will converge to a fixed point. Convergence can be improved by using damping in Equations 7.

Here, 2 is an identifier of the formula, $b$ is an identifier of the atom occurrence in the formula, and $<X,Y>$ are the logical variables appearing in the formula.

As in the ground case, each first-order cloning is associated with a corresponding equivalence between the original atom and its clone, except that the equivalence is first-order in this case. The first-order cloning of atom occurrence $\text{smokes}(Y)$ into $\text{smokes}_{2b<X,Y>}(Y)$ in the example above leads to introducing the following first-order equivalence:

$$\text{smokes}(Y) \Leftrightarrow \text{smokes}_{2b<X,Y>}(Y) \qquad (9)$$

Let us now consider the groundings of Formulas 8 and 9, assuming a domain of $\{a,b\}$:

1.5    $\text{smokes}(a) \wedge \text{friends}(a,a) \Rightarrow \text{smokes}_{2b<a,a>}(a)$
1.5    $\text{smokes}(a) \wedge \text{friends}(a,b) \Rightarrow \text{smokes}_{2b<a,b>}(b)$
1.5    $\text{smokes}(b) \wedge \text{friends}(b,a) \Rightarrow \text{smokes}_{2b<b,a>}(a)$
1.5    $\text{smokes}(b) \wedge \text{friends}(b,b) \Rightarrow \text{smokes}_{2b<b,b>}(b)$
       $\text{smokes}(a) \Leftrightarrow \text{smokes}_{2b<a,a>}(a)$
       $\text{smokes}(b) \Leftrightarrow \text{smokes}_{2b<a,b>}(b)$
       $\text{smokes}(a) \Leftrightarrow \text{smokes}_{2b<b,a>}(a)$
       $\text{smokes}(b) \Leftrightarrow \text{smokes}_{2b<b,b>}(b)$

We have a few observations on the proposed cloning and relaxation techniques. First, the four groundings of (8) contain distinct groundings of the clone $\text{smokes}_{2b<X,Y>}(Y)$. Second, if we relax the equivalence in (9), the ground formulas of (8) will no longer interact through the clone $\text{smokes}_{2b<X,Y>}(Y)$. Third, if we did not append the logical variables $<X,Y>$ during the cloning process, the previous statement would no longer hold. In particular, without appending logical variables, the four groundings of (8) would have contained only the two distinct clone groundings, $\text{smokes}_{2b}(a)$ and $\text{smokes}_{2b}(b)$. This will lead to continued interactions between the four groundings of (8).[4]

The proposed cloning technique leads to MLNs in which one quantifies over predicate names (as in second-order logic). This can be avoided, but it leads to less transparent semantics. In particular, we can avoid quantifying over predicate names by using ground predicate names with increased arity. For example, $\text{smokes}_{2b<X,Y>}(Y)$ could have been written as $\text{smokes}_{2b}(X,Y)$ where we pushed $<X,Y>$ into the predicate arguments. The disadvantage of this, however, is that the semantics of the individual arguments is lost as the arguments become overloaded.

---

[4]We need to remove all interactions among groundings of the same formula because otherwise the formula might not even be tractable for exact inference. For example, there is currently no exact lifted inference algorithm that can handle the formula $w: \quad \text{p}_a(X,Y) \wedge \text{p}_b(Y,Z) \Rightarrow \text{p}_c(X,Z)$ without grounding it first (Van den Broeck, 2011).

We now have the following key theorem.

**Theorem 1.** *Let $\Delta^r$ be the MLN resulting from cloning all atom occurrences in MLN $\Delta$ and then relaxing all introduced equivalences. Let $\Delta^g$ be the grounding of $\Delta^r$. The formulas of $\Delta^g$ are then fully disconnected (i.e., they share no atoms).*

With this theorem, the proposed first-order cloning and relaxation technique allows one to fully disconnect the grounding of an MLN by simply relaxing first-order equivalences in the first-order MLN.

### 3.2 FIRST-ORDER COMPENSATION

In principle, one can just clone atom occurrences, relax some equivalence constraints, and then use the resulting MLN as an approximation of the original MLN. By relaxing enough equivalences, the approximate MLN can be made arbitrarily easy for exact inference. Our goal in this section, however, is to improve the quality of approximations by compensating for the relaxed equivalences, yet without making the relaxed MLN any harder for exact inference. This will be done through a notion of first-order compensation.

#### 3.2.1 Equiprobable Equivalences

The proposed technique is similar to the one for ground MLNs, that is, using *weighted atoms* whose weights will allow for compensation. The key, however, is to use first-order weighted atoms instead of ground ones. For this, we need to define the following notions.

**Definition 2** (Equiprobable Set). A set of random variables $V$ is called equiprobable w.r.t. distribution Pr iff for all $v_1, v_2 \in V : \Pr(v_1) = \Pr(v_2)$.

**Definition 3** (Equiprobable Equivalence). Let $\Delta$ be an MLN from which a first-order equivalence $a \Leftrightarrow a'$ was relaxed. Let $a_1 \Leftrightarrow a'_1, \ldots, a_n \Leftrightarrow a'_n$ be all groundings of $a \Leftrightarrow a'$. The equivalence $a \Leftrightarrow a'$ is equiprobable iff the sets $\{a_1, \ldots, a_n\}$ and $\{a'_1, \ldots, a'_n\}$ are both equiprobable w.r.t the ground distribution of MLN $\Delta$.

The basic idea of first-order compensation is that when relaxing an equiprobable equivalence $a \Leftrightarrow a'$, under certain conditions, one can compensate for its loss using only two weighted first-order atoms of the form:

$$w: \quad a \quad \text{and} \quad w': \quad a'$$

This follows because if we were to fully ground the equivalence into $a_1 \Leftrightarrow a'_1, \ldots, a_n \Leftrightarrow a'_n$ and then apply ground compensation, the relevant ground atoms will attain the same weights. That is, by the end of ground compensation, the weighted ground atoms,

$$w_i: \quad a_i \quad \text{and} \quad w'_i: \quad a'_i$$

will have $w_i = w_j$ and $w'_i = w'_j$ for all $i$ and $j$.

### 3.2.2 Partitioning Equivalences

To realize first-order compensation, one must address two issues. First, a relaxed first-order equivalence may not be equiprobable to start with. Second, even when the equivalence is equiprobable, it may cease to be equiprobable as we adjust the weights during the compensation process. Recall that equiprobability is defined with respect to the ground distribution of an MLN. Yet, this distribution changes during the compensation process, which iteratively changes the weights of compensating atoms and, hence, also iteratively changes the ground distribution.

Consider for example the following relaxed equivalences: $p(X) \Leftrightarrow q(X)$ and $q(X) \Leftrightarrow r(X)$. Suppose the domain is $\{a, b\}$ and the current ground distribution, $\Pr_i$, is such that $\Pr_i(p(a)) = \Pr_i(p(b))$, $\Pr_i(q(a)) = \Pr_i(q(b))$, and $\Pr_i(r(a)) \neq \Pr_i(r(b))$. In this case, the equivalence $p(X) \Leftrightarrow q(X)$ is equiprobable, but $q(X) \Leftrightarrow r(X)$ is not equiprobable.

If an equivalence constraint is not equiprobable, one can always partition it into a set of equiprobable equivalences — in the worst case, the partition will include all groundings of the equivalence. In the above example, one can partition the equivalence $q(X) \Leftrightarrow r(X)$ into the equivalences $q(a) \Leftrightarrow r(a)$ and $q(b) \Leftrightarrow r(b)$, which are trivially equiprobable.

Given this partitioning, the compensation algorithm will add distinct weights for the compensating atoms $q(a)$ and $q(b)$. Therefore, the set $\{q(a), q(b)\}$ will no longer be equiprobable in the next ground distribution, $\Pr_{i+1}$. As a result, the equivalence $p(X) \Leftrightarrow q(X)$ will no longer be equiprobable w.r.t. the ground distribution $\Pr_{i+1}$, even though it was equiprobable with respect to the previous ground distribution $\Pr_i$.

### 3.2.3 Strongly Equiprobable Equivalences

To attain the highest degree of lifting during compensation, one needs to dynamically partition equivalences after each iteration of the compensation algorithm, to ensure equiprobability. We defer the discussion on dynamic partitioning to Appendix A, focusing here on a strong version of equiprobability that allows one to circumvent the need for dynamic partitioning.

The mentioned technique is employed by our current implementation of Lifted RCR, which starts with equivalences that are *strongly equiprobable.* An equivalence is strongly equiprobable if it is equiprobable w.r.t all ground distributions induced by the compensation algorithm (i.e., ground distributions that result from only modifying the weights of compensating atoms).

Consider again Formula 2 where we cloned the atom occurrence $\text{smokes}(Y)$ and relaxed its equivalence, leading to the MLN:

1.5 $\text{smokes}(X) \land \text{friends}(X, Y) \Rightarrow \text{smokes}_{2b<X,Y>}(Y)$

and relaxed equivalence

$$\text{smokes}(Y) \Leftrightarrow \text{smokes}_{2b<X,Y>}(Y) \qquad (10)$$

Suppose we partition this equivalence as follows:[5]

$$X = Y, \ \text{smokes}(Y) \Leftrightarrow \text{smokes}_{2b<X,Y>}(Y)$$
$$X \neq Y, \ \text{smokes}(Y) \Leftrightarrow \text{smokes}_{2b<X,Y>}(Y)$$

These equivalences are not only equiprobable w.r.t. the relaxed MLN, but also strongly equiprobable. That is, suppose we add to the relaxed model the compensating atoms

$$w_1 : \quad \text{smokes}(X)$$
$$w'_1 : \ X = Y, \ \text{smokes}_{2b<X,Y>}(Y)$$
$$w_2 : \quad \text{smokes}(X)$$
$$w'_2 : \ X \neq Y, \ \text{smokes}_{2b<X,Y>}(Y)$$

The two equivalences will be equiprobable w.r.t. any ground distribution that results from adjusting the weights of these compensating atoms. We will present an equivalence partitioning algorithm in Section 4 that guarantees strong equiprobability of the partitioned equivalences. This algorithm is employed by our current implementation of Lifted RCR and will be used when reporting experimental results later.

### 3.3 COUNT-NORMALIZATION

We are one step away from presenting our first-order compensation scheme. What is still missing is a discussion of *count-normalized* equivalences, which are also required by our compensation scheme.

Consider Equivalence 10, which has four groundings

$$\text{smokes}(a) \Leftrightarrow \text{smokes}_{2b<a,a>}(a)$$
$$\text{smokes}(b) \Leftrightarrow \text{smokes}_{2b<a,b>}(b)$$
$$\text{smokes}(a) \Leftrightarrow \text{smokes}_{2b<b,a>}(a)$$
$$\text{smokes}(b) \Leftrightarrow \text{smokes}_{2b<b,b>}(b)$$

for the domain $\{a, b\}$. There are two distinct groundings of the original atom $\text{smokes}(Y)$ in this case and each of them appears in two groundings. When each grounding of the original atom appears in exactly the same number of ground equivalences, we say that the first-order equivalence is count-normalized.

---
[5]We are using an extension of MLNs that allows constraints, such as $X \neq Y$. Our implementation is in terms of parfactor graphs, which are more expressive than standard MLNs and do allow for the representation of such constraints. In extended MLNs, we will write $C, f$ to mean that $C$ is a constraint that applies to formula $f$.

Consider now a constrained version of Equivalence 10

$$X \neq b \vee Y \neq b, \ \text{smokes}(Y) \Leftrightarrow \text{smokes}_{2b<X,Y>}(Y)$$

which has the following groundings

$$\text{smokes}(a) \Leftrightarrow \text{smokes}_{2b<a,a>}(a)$$
$$\text{smokes}(b) \Leftrightarrow \text{smokes}_{2b<a,b>}(b)$$
$$\text{smokes}(a) \Leftrightarrow \text{smokes}_{2b<b,a>}(a)$$

This constrained equivalence is not count-normalized since the atom $\text{smokes}(a)$ appears in two ground equivalences while the atom $\text{smokes}(b)$ appears in only one. More generally, we have the following definition.

**Definition 4.** Let $C, \ a \Leftrightarrow a'$ be a first-order equivalence. Let $\alpha$ be an instantiation of the variables in original atom $a$ and assume that $\alpha$ satisfies constraint $C$. The equivalence is *count-normalized* iff $C \wedge \alpha$ has the same number of solutions for each instantiation $\alpha$. Moreover, the number of groundings for $C, \ a$ is called the *original count* and the number of groundings for $C, \ a'$ is called the *clone count*.

Count-normalization can only be violated by constrained equivalences. Moreover, for a certain class of constraints, count-normalization is always preserved. The algorithm we shall present in Section 4 for partitioning equivalences takes advantage of this observation. In particular, the algorithm generates constrained equivalences whose constraint structure guarantees count-normalization.

### 3.4 THE COMPENSATION SCHEME

We now have the following theorem.

**Theorem 2.** *Let $\Delta_i$ be an MLN with relaxed equivalences $C, \ a \Leftrightarrow a'$ and, hence, corresponding compensating atoms:*

$$w_i: \quad C, \ a \qquad \text{and} \qquad w'_i: \quad C, \ a'$$

*Suppose that the equivalences are count-normalized and strongly equiprobable. Let $a_g \Leftrightarrow a'_g$ be one grounding of equivalence $C, \ a \Leftrightarrow a'$, let $n$ be its original count and $n'$ be its clone count. Consider now the MLN $\Delta_{i+1}$ obtained using the following updates:*

$$w_{i+1} = \frac{n'}{n} \left( \log\left(\Pr{}_i(a'_g)\right) - \log\left(\Pr{}_i(\neg a'_g)\right) - w'_i \right)$$
$$w'_{i+1} = \log\left(\Pr{}_i(a_g)\right) - \log\left(\Pr{}_i(\neg a_g)\right) - w_i \qquad (11)$$

*The ground distribution of MLN $\Delta_{i+1}$ equals the one obtained by applying Ground RCR to MLN $\Delta_i$.*

Note that first-order compensation requires exact inference on the MLN $\Delta_i$, which is needed for computing $\Pr_i(a_g)$ and $\Pr_i(a'_g)$. Moreover, these computations will need to be repeated until one obtains a fixed point of the update equations given by Theorem 2.

### 3.5 FIRST-ORDER RECOVERY

Recovering a first-order equivalence $C, \ a \Leftrightarrow a'$ amounts to removing its compensating atoms

$$w_i: \quad C, \ a \qquad \text{and} \qquad w'_i: \quad C, \ a'$$

and then adding the equivalence back to the MLN.

Adapting the ground recovery heuristic suggested earlier, one recovers the first-order equivalence that maximizes the symmetric pairwise KL-divergence

$$n' \cdot \text{KLD}\left( \Pr(a_g), \Pr(a'_g), \frac{e^{w_i+w'_i}}{1+e^{w_i+w'_i}} \right),$$

where $n'$ is the clone count of the equivalence. Note here that $n'$ is also the number of equivalence groundings since, by definition, the clone atom contains all logical variables that appear in the equivalence.

Note that recovering first-order equivalences may violate the equiprobability of equivalences that remain relaxed, which in turn may require re-partitioning.

## 4 PARTITIONING EQUIVALENCES

We will now discuss a method for partitioning first-order equivalences, which guarantees both strong equiprobability and count-normalization. This method is used by our current implementation of Lifted RCR that we describe in Section 6.

Our method is based on the procedure of *preemptive shattering* given by Poole et al. (2011), which is a conceptually simpler version of the influential *shattering* algorithm proposed by Poole (2003) and de Salvo Braz et al. (2005) in the context of exact lifted inference. We will first describe this shattering procedure, which partitions atoms. We will then use it to partition all atoms in a relaxed MLN. We will finally show how these atom partitions can be used to partition first-order equivalences.

### 4.1 PREEMPTIVE SHATTERING

Preemptive shattering takes as input an atom $p(X_1, \ldots, X_n)$ and a set of constants $K = \{k_1, \ldots, k_m\}$. It then returns a set of constrained atoms of the form $C, \ p(X_1, \ldots, X_n)$ which represent a partitioning of the input atom. That is, the groundings of constrained atoms are guaranteed to be disjoint and cover all groundings of the input atom.

We start with an intuitive description of preemptive shattering. The set of constants $K$ are the ones explicitly mentioned in the MLN of interest. If an argument $X_i$ of the input atom $p(X_1, \ldots, X_n)$ can take on one of the constants $k_j \in K$, preemptive shattering splits the atom into two constrained atoms: one where $X_i$ is substituted by constant $k_i$ and one where the constant $k_j$ is excluded from the domain of $X_i$. Moreover, when two arguments $X_i$ and $X_j$ of the input atom can take on the same value, preemptive shattering splits the atom into two constrained atoms: one with the constraint $X_i = X_j$ and another with $X_i \neq X_j$.

We will next describe the shattering procedure and its complexity more formally. We need some definitions first. For each argument $X_i$ of the input atom, let

$$C_i = \{(X_i = k_1), \ldots, (X_i = k_m),\\ (X_i \neq k_1, \ldots, X_i \neq k_m)\}.$$

Preemptive shattering generates all possible combinations of the above constraints:

$$\mathbf{C}_A = \{c_1 \wedge \cdots \wedge c_n \mid (c_1, \ldots, c_n) \in \times_{i=1}^n C_i\},$$

where $\times_{i=1}^n C_i$ is the cartesian product of $C_1, \ldots, C_n$.

Consider now a subset $\mathcal{X} \subseteq \{X_1, \ldots, X_n\}$ of the input atom arguments and let

$$\mathbf{C}(\mathcal{X}) = \bigwedge_{X_i, X_j \in \mathcal{X}} (X_i = X_j) \wedge \bigwedge_{i=1}^n \bigwedge_{X_j \notin \mathcal{X}} (X_i \neq X_j)$$

$$\mathbf{C}_B = \{\mathbf{C}(\mathcal{X}) \mid \mathcal{X} \subseteq \{X_1, \ldots, X_n\}\}$$

Preemptive shattering will then create the following partition of the input atom $p(X_1, \ldots, X_n)$:

$$A = \{c_a \wedge c_b, \ p(X_1, \ldots, X_n) \mid c_a \in \mathbf{C}_A, c_b \in \mathbf{C}_B\}$$

Many of the generated constraints $c_a \wedge c_b$ will have no solutions and can be dropped. What remains is a set of constrained atoms whose groundings form a partition of the groundings for the input atom. Preemptive shattering can be implemented in time that is exponential in the arity $n$ of input atom and polynomial in the number of input constants $m$.

For an example, consider the formula $\text{smokes}(X) \Leftrightarrow \text{smokes}_{<X,Y>}(X)$ and assume we have evidence $\text{smokes}(a)$ and therefore $K = \{a\}$. Preemptive shattering of the input atom $\text{smokes}(X)$ returns back

$$X = a, \ \text{smokes}(X)$$
$$X \neq a, \ \text{smokes}(X)$$

Preemptive shattering of the input atom $\text{smokes}_{<X,Y>}(X)$ returns back

$$X = a, Y = a, \ \text{smokes}_{<X,Y>}(X)$$
$$X = a, Y \neq a, \ \text{smokes}_{<X,Y>}(X)$$
$$X \neq a, Y = a, \ \text{smokes}_{<X,Y>}(X)$$
$$X \neq a, Y \neq a, X = Y, \ \text{smokes}_{<X,Y>}(X)$$
$$X \neq a, Y \neq a, X \neq Y, \ \text{smokes}_{<X,Y>}(X)$$

We will next show how this shattering procedure forms the basis of a method for partitioning equivalence constraints, with the aim of ensuring both strong equiprobability and count-normalization.

### 4.2 PARTITIONING EQUIVALENCES BY PREEMPTIVE SHATTERING

Consider an MLN which results from cloning some atom occurrences and then adding corresponding equivalence constraints. Let $K$ be all the constants appearing explicitly in the MLN.

To partition a first-order equivalence $a \Leftrightarrow a'$, our method will first apply preemptive shattering to the original atom $a$ and clone atom $a'$, yielding a partition for each. Suppose that $C_1, a_1, \ldots, C_n, a_n$ is the partition returned for original atom $a$. Suppose further that $C'_1, a'_1, \ldots, C'_m, a'_m$ is the partition returned for clone atom $a'$. By definition of cloning, all variables that appear in original atom $a$ must also appear in clone atom $a'$. This implies the following property. For every (original) constraint $C_i$, there is a corresponding set of (clone) constraints $C'_j$ that partition $C_i$. Each pair of constraints $C_i$ and $C'_j$ will then generate a member of the equivalence partition: $C_i \wedge C'_j, a_i \Leftrightarrow a'_j$. Note that $C_i \wedge C'_j = C'_j$ since $C'_j$ implies $C_i$.

Let us consider an example. Suppose we are partitioning the equivalence $\text{smokes}(X) \Leftrightarrow \text{smokes}_{<X,Y>}(X)$. Section 4.1 showed the preemptive shattering of the atoms $\text{smokes}(X)$ and $\text{smokes}_{<X,Y>}(X)$. These give rise to the following equivalence partition:

$$X = a, Y = a, \text{smokes}(X) \Leftrightarrow \text{smokes}_{<X,Y>}(X)$$
$$X = a, Y \neq a, \text{smokes}(X) \Leftrightarrow \text{smokes}_{<X,Y>}(X)$$
$$X \neq a, Y = a, \text{smokes}(X) \Leftrightarrow \text{smokes}_{<X,Y>}(X)$$
$$X \neq a, Y \neq a, X = Y, \text{smokes}(X) \Leftrightarrow \text{smokes}_{<X,Y>}(X)$$
$$X \neq a, Y \neq a, X \neq Y, \text{smokes}(X) \Leftrightarrow \text{smokes}_{<X,Y>}(X)$$

The following result is proven in Appendix B.

**Theorem 3.** *Partitioning by preemptive shattering returns count-normalized, strongly equiprobable equivalences.*

When $K$ is small, preemptive shattering will find partitions that are close to minimal. When $K$ is large, however, it will create large partitions, defeating the purpose of lifted inference. In the next section, we will mention some alternative partitioning algorithms that work on a fully relaxed model. However, preemptive shattering is the only partitioning algorithm to our knowledge that works for any level of relaxation. We believe our work can motivate future work on finding more efficient general partitioning algorithms and even approximate partitioning algorithms.

## 5 RELATED WORK

The RCR framework has previously been used to characterize iterative belief propagation (IBP) and some of its generalizations. In the case where the simplified model is fully disconnected, the approximate marginals of RCR correspond to the approximate marginals given by IBP (Pearl, 1988; Choi and Darwiche, 2006). The approximation to the partition function further corresponds to the *Bethe free energy* approximation (Yedidia et al., 2003; Choi and Darwiche, 2008). When equivalence constraints have been recovered, RCR corresponds to a class of *generalized belief propagation* (GBP) approximations (Yedidia et al., 2003), and in particular *iterative joingraph propagation* (Aji and McEliece, 2001; Dechter et al., 2002). RCR also inspired a system that was successfully employed in a recent approximate inference competition (Elidan and Globerson, 2010; Choi and Darwiche, 2011). *Mini-buckets* can also be viewed as an instance of RCR where no compensations are used (Kask and Dechter, 2001; Dechter and Rish, 2003; Choi et al., 2007), which leads to upper bounds on the partition function (for example). Any approximate MLN found by Lifted RCR corresponds to one found by Ground RCR on the ground MLN, thus all of the above results carry over to the lifted setting.

The motivation for calling our approach *lifted* is threefold. First, in the compensation phase, we are compensating for many ground equivalences at the same time. Computing compensating weights for all of these requires inferring only a *single* pair of marginal probabilities. Second, computing marginal probabilities is done by an *exact lifted* inference algorithm. Third, we relax and recover first-order equivalence constraints, which correspond to *sets* of ground equivalences.

The work on lifted approximate inference has mainly focused on lifting the IBP algorithm. The correspondence between IBP and RCR carries over to their lifted counterparts: Lifted RCR compensations on a fully relaxed model correspond to *lifted belief propagation* (Singla and Domingos, 2008). Starting from a first-order model, Singla and Domingos (2008) proposed *lifted network construction* (LNC), which partitions atoms into so-called *supernodes*. The ground atoms represented by these supernodes send and receive the same messages when running IBP. This means that they partition the atoms into equiprobable sets and that LNC can be used for equivalence partitioning in Lifted RCR for the fully relaxed model. Kersting et al. (2009) proposed a *color-passing* (CP) algorithm that achieves similar results as LNC, only starting from a ground model, where the first-order structure is not apparent. Two other approximate lifted inference algorithms are *probabilistic theorem proving* (Gogate and Domingos, 2011), which contains a Monte-Carlo method and *bisimulation-based lifted inference* (Sen et al., 2009), which uses a mini-bucket approximation on a model that was compressed by detecting symmetries. Because of the correspondence between Ground RCR and mini-buckets mentioned above, this approach can also be seen as an instance of Lifted RCR with the compensation phase removed.

## 6 EXPERIMENTS

In this section, we evaluate the Lifted RCR algorithm on common benchmarks from the lifted inference literature. The experiments were set up to answer the questions: **(Q1)** To which extent does recovering first-order equivalences improve the approximations found by Lifted RCR? **(Q2)** Can IBP be improved considerably through the recovery of a small number of equivalences? **(Q3)** Is there a significant advantage to using Lifted RCR over Ground RCR?

We implemented Lifted RCR and released it as open source software.[6] To compute exact marginal probabilities in Equations 11, we use first-order knowledge compilation (Van den Broeck et al., 2011). It compiles the MLN into a logical circuit where probabilistic inference is performed by weighted model counting, which exploits context-specific independencies and determinism in the MLN. It is arguably the state of the art in exact lifted inference (Van den Broeck, 2011; Van den Broeck and Davis, 2012). In combination with preemptive shattering, we compile a first-order circuit once and re-evaluate it in each iteration of the compensation algorithm. This is possible because the structure of the compensated MLNs do not change between iterations, only their parameters change. An already compiled first-order circuit can re-evaluated very efficiently. See Appendix C for further details.

To answer **(Q1-2)** we ran Lifted RCR on MLNs from the exact lifted inference literature, where computing exact marginals is tractable. This allows us to evaluate the approximation quality of Lifted RCR for different degrees of relaxation. We used the models *p-r* and *sick-death* (de Salvo Braz et al., 2005), *work-*

---

[6] http://dtai.cs.kuleuven.be/ml/systems/wfomc

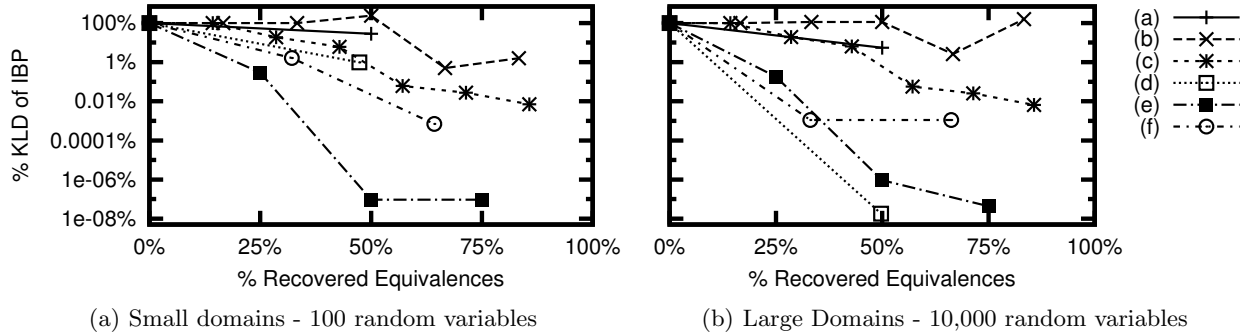

(a) Small domains - 100 random variables

(b) Large Domains - 10,000 random variables

Figure 1: Normalized approximation error of Lifted RCR for different levels of approximation on the models (a) *p-r*, (b) *sick-death*, (c) *workshop attributes*, (d) *smokers*, (e) *smokers and drinkers* and (f) *symmetric smokers*.

*shop attributes* (Milch et al., 2008), *smokers* (Singla and Domingos, 2008), *smokers and drinkers* (Van den Broeck et al., 2011) and *symmetric smokers* (Van den Broeck, 2011). Each of these models represents a new advance in exact lifted inference. They are incrementally more complex and challenging for lifted inference. The results are shown in Figure 1, where we ran Lifted RCR on the above models for two sets of domain sizes: a small and a large set, where the number of random variables is on the order of 100 and 10,000 respectively. We plot the symmetric KL divergence between the exact marginals and the approximations found by Lifted RCR, as a percentage of the KL divergence of the fully relaxed approximation. The horizontal axis shows the level of relaxation in terms of the percentage of recovered ground equivalences. The 0% point corresponds to the approximations found by (lifted) IBP. The 100% point corresponds to exact inference.[7]

We see that each recovered first-order equivalence tends to improve the approximation quality significantly, often by more than one order of magnitude, answering **(Q1)**. In the case of *smokers* with a large domain size, recovering a single equivalence more than the IBP approximation reduced the KL divergence by 10 orders of magnitude, giving a positive answer to **(Q2)**. The *sick-death* model is the only negative case for **(Q2)**, where recovering equivalences does not lead to approximations that are better than IBP.[8]

To answer **(Q3)**, first note that as argued in Section 5, the 0% recovery point of Lifted RCR using LNC or CP to partition equivalences corresponds to lifted IBP. For this case, the work of Singla and Domingos (2008) and Kersting et al. (2009) has extensively shown that Lifted IBP/RCR methods can significantly outperform Ground IBP/RCR. Similarly, computational gains for the 100% recovery point were shown in the exact lifted inference literature. For intermediate levels of relaxation, we ran Ground RCR on the above models with large domain sizes. On these, Ground RCR could never recover more than 5% of the relaxed equivalences before exact inference in the relaxed model becomes intractable. This answers **(Q3)** positively.

For the above experiments, Appendix D further reports on the quality of the approximations and the convergence of the compensation algorithm, both as a function of runtime and the number of iterations.

## 7 CONCLUSIONS

We presented Lifted RCR, a lifted approximate inference algorithm that performs exact inference in a simplified model. We showed how to obtain a simplified model by relaxing first-order equivalences, compensating for their loss, and recovering them as long as exact inference remains tractable. The algorithm can traverse an entire spectrum of approximations, from lifted iterative belief propagation to exact lifted inference. Inside this spectrum is a family of lifted joingraph propagation (and GBP) approximations. We empirically showed that recovering first-order equivalences in a relaxed model can substantially improve the quality of an approximation. We also remark that Lifted RCR relies on an exact lifted inference engine as a black box, and that any future advances in exact lifted inference have immediate impact in lifted approximate inference.

### Acknowledgements

This work has been partially supported by ONR grant #N00014-12-1-0423, NSF grant #IIS-1118122, and NSF grant #IIS-0916161. GVdB is supported by the Research Foundation-Flanders (FWO-Vlaanderen).

---

[7] All experiments ran up to the 100% point, which is not shown in the plot because it has a KL divergence of 0.

[8] Interestingly, it is also the only example where some compensations failed to converge without using damping.